\documentclass{article}
\usepackage[T1]{fontenc}
\usepackage[utf8]{inputenc}
\usepackage{microtype}
\usepackage{graphicx}
\usepackage{booktabs}
\usepackage{amsmath,amssymb}
\usepackage{array,tabularx}

\usepackage{multirow}

\usepackage[accepted]{mlsys2025}
\usepackage{etoolbox}
\makeatletter
\patchcmd{\printAffiliationsAndNotice}
{\ \\ \Notice@String}
{}
{}
{\PackageError{paper}{Failed to remove notice}{}}
\makeatother
\usepackage{xurl}
\usepackage[hyperfootnotes=false]{hyperref}
\newcommand{\stepnum}[1]{\textnormal{\textcircled{\scriptsize #1}}}
\providecommand{\tightlist}{\setlength{\itemsep}{0pt}\setlength{\parskip}{0pt}}

\DeclareUrlCommand\nolinkurl{\urlstyle{same}}
\definecolor{paperlink}{rgb}{0,0.08,0.45}
\hypersetup{colorlinks=true,allcolors=paperlink,pdftitle={FlashOffloading: Low-Latency KV Cache Offloading for Long-Context LLM Serving},
 pdfauthor={Anonymous Submission},pdfsubject={Long-context LLM inference and KV cache offloading}}
\mlsystitlerunning{FlashOffloading: Low-Latency KV Cache Offloading for Long-Context LLM Serving}
\AtBeginDocument{%
	\setlength{\abovedisplayskip}{3pt plus 1pt minus 1pt}
	\setlength{\belowdisplayskip}{3pt plus 1pt minus 1pt}
	\setlength{\abovedisplayshortskip}{0pt plus 1pt}
	\setlength{\belowdisplayshortskip}{3pt plus 1pt minus 1pt}
}
\begin{document}
\twocolumn[
\mlsystitle{FlashOffloading: Low-Latency KV Cache Offloading for Long-Context LLM Serving}
\begin{mlsysauthorlist}
\mlsysauthor{Xuan Wang}{anon}
\mlsysauthor{Zhuanhao Wu}{anon}
\mlsysauthor{Tingzhu Fang}{anon,sec}
\mlsysauthor{Xushi Li}{anon,sec}
\mlsysauthor{Sicheng Zhang}{anon,sec}
\mlsysauthor{ZHENYU YANG}{anon}
\mlsysauthor{Feng Liu}{anon,sec}
\mlsysauthor{Zhaohui Fan}{anon}
\mlsysauthor{Qiang Xie}{anon}
\mlsysauthor{Huayi Zou}{anon}
\mlsysauthor{Yuncong Liu}{anon,sec}
\mlsysauthor{Zeyu Song}{anon,sec}
\mlsysauthor{Ke Zhu}{anon,sec}
\mlsysauthor{Yanjia Zeng}{anon,sec}
\mlsysauthor{Jingchun Gao}{anon,sec}
\mlsysauthor{Yan Shi}{anon,third}
\mlsysauthor{Kui Luo}{anon,sec}
\mlsysauthor{Gezi Li}{anon}
\mlsysauthor{LiangjunFeng}{anon,sec}
\mlsysauthor{Cheng Li}{anon}
\end{mlsysauthorlist}
\mlsysaffiliation{anon}{Huawei Technologies Co., Ltd}
\mlsysaffiliation{sec}{Huawei Ascend Computing Technology Development}
\mlsysaffiliation{third}{Shanghai Jiao Tong University}
\mlsyscorrespondingauthor{Cheng Li}{licheng56@huawei.com}
\mlsyskeywords{LLM inference, KV cache offloading, sparse attention, long context}
\vskip 0.3in
\begin{abstract}
In large language model (LLM) serving systems, the key-value (KV) cache grows linearly with sequence length and concurrency, so long-context inference easily hits the high-bandwidth memory (HBM) capacity wall and results in low concurrency.
Offloading the KV cache to host dynamic random-access memory (DRAM) frees HBM space and raises concurrency, but the additional cache management and host-to-device (H2D) load overhead inflates inference latency,
higher concurrency fails to translate into effective throughput.
We present FlashOffloading, a host-DRAM-centric inference architecture that fully offloads the KV cache in both the prefill and decode phases.
FlashOffloading builds rolling HBM cache for prefill, a hot cache for decode, a reverse sparse index with zero extra miss-lookup overhead, and a compute--load overlapped kernel, optimizing KV offloading latency across the entire pipeline and delivering throughput gains under a time-per-output-token (TPOT) service-level objective (SLO).
In experiments deploying GLM-5.2 with multi-token prediction (MTP3) speculative decoding on an Ascend910C prefill--decode (PD)-disaggregated platform, FlashOffloading improves prefill throughput by up to 1.38\(\times\) and decode throughput by up to 1.91\(\times\) under a 30 ms TPOT constraint.
FlashOffloading is integrated into an open-source serving framework: https://github.com/vllm-project/vllm-ascend/tree/releases/v0.30.0.
\end{abstract}
]
\begingroup

\printAffiliationsAndNotice{}
\endgroup
\section{Introduction}\label{sec:introduction}

Long-document understanding, repository-scale code analysis, and long-horizon agent interactions are driving LLM context windows toward millions of tokens. Mainstream online serving systems widely adopt PD disaggregation: P nodes perform chunked prefill and D nodes autoregressive decoding. To avoid recomputation, both stages retain full keys and values for every historical token and attention layer (KV cache), creating substantial memory pressure; a typical 1M-token sequence requires 10--100 GB of KV cache. With finite HBM on graphics processing units (GPUs) and neural processing units (NPUs), this limits concurrency and throughput and can make ultra-long requests unservable.

Tiered KV cache offloading extends available memory with host DRAM or storage, offering a direct way around the HBM capacity wall \cite{ref6,ref7,ref28}. However, these devices provide much lower bandwidth than HBM, so early work targets offline inference or merely makes long sequences servable. FlexGen achieves high-throughput, latency-insensitive inference through layerwise KV cache offloading and loading \cite{ref2}; subsequent systems stream KV state \cite{ref3} or offload attention computation and state to tiered memory \cite{ref8,ref5,ref27}, enabling longer sequences and larger batches. Nevertheless, repeated KV transfers still incur substantial bandwidth and latency overheads for online serving \cite{ref22,ref23}.

Reducing transfer volume is therefore essential for online tiered-memory serving. Prior work exploits attention sparsity and approximate selection \cite{ref9,ref10,ref34,ref31,ref24,ref29} to retain only a small set of KV cache blocks for attention. ShadowKV uses low-rank key representations \cite{ref11}, FreeKV employs speculative retrieval and correction \cite{ref12}, and SparseServe and SAC coordinate hierarchical cache management \cite{ref13,ref14}. These methods reduce bandwidth demand but often degrade model quality by at least 1\%, limiting their deployment in real online systems.

Native sparse attention, exemplified by DeepSeek Sparse Attention (DSA), changes this premise. Models such as DeepSeek-V3.2 and GLM-5.2 use learned indexers to select top-\(k\) KV entries \cite{ref15,ref16}: the full history must be retained, but each decode step accesses only a small, fine-grained KV cache subset. ESS \cite{ref33}, HiSparse \cite{ref30}, ECHO \cite{ref20}, and AVSG \cite{ref35} therefore use HBM--DRAM two-level caches to reduce load volume and offload decode KV cache without degrading model quality. However, cache indexing and fine-grained host-to-device (H2D) loads still substantially increase inference latency.

We observe that prefill and decode differ markedly in computation and KV cache access. Prefill offers ample layer-computation windows and coarse accesses, making bulk transfers easier to overlap with computation. Decode needs only indexer-selected entries: the total H2D volume is small, but accesses are fine-grained (\(\sim\)1 KB) and cannot start until indexing completes, making them difficult to overlap with computation.

Motivated by this observation, we present FlashOffloading, an architecture that offloads the KV cache in both prefill and decode for real-world LLM serving. For prefill, a pool of four layer buffers keeps the first layer resident and rotates the other three buffers across subsequent layers, overlapping H2D loads with layer computation. For decode, a reverse sparse index supports a large hot cache to reduce misses and load volume; miss-lookups are fused into top-\(k\) to eliminate extra miss-lookup latency; and fine-grained fused kernels overlap H2D loads with sparse attention.

This work makes the following contributions:

\begin{enumerate}
\def\labelenumi{\arabic{enumi}.}
\tightlist
\item
  \textbf{A phase-specialized host-backed KV architecture.} We combine layerwise prefill offloading and exact token-selective decode retrieval under a unified host-backed KV architecture.
\item
  \textbf{A bounded-memory layerwise pipeline for prefill.} A resident first-layer buffer and three rotating buffers overlap loading, computation, and writeback to overlap transfers with layer computation.
\item
  \textbf{A low-latency, exact decode-offloading engine.} A reverse sparse index, cache-management kernels with no additional miss-lookup overhead, and compute--load overlap kernels substantially reduce the added cache management and H2D-load latency.
\item
  \textbf{An end-to-end implementation and evaluation.} We integrate both paths into vLLM-Ascend and deploy GLM-5.2 with MTP3 speculative decoding on an Ascend910C PD-disaggregated platform, improving prefill throughput by up to 1.38\(\times\) and decode throughput by up to 1.91\(\times\) under a 30 ms TPOT constraint.
\end{enumerate}

\section{Background and Motivation}\label{sec:background}

\subsection{The KV Cache Capacity Problem}\label{sec:kv-capacity}

Long-sequence workloads create enormous KV cache capacity demands, challenging the performance of LLM serving systems. In LLM inference, the prefill phase processes the input sequence in chunked forward passes and generates the KV cache; the decode phase attends over the full historical KV cache at every step to emit a new token (or multiple tokens under speculative decoding). After deducting the space for weights and activations, mainstream inference frameworks \cite{ref1} pre-allocate the remaining HBM to the KV cache, requiring the total KV of all running requests to fit within it:

\begin{equation}
\label{eq:kvalloc}
\begin{gathered}
M_{\mathrm{KValloc}} = M_{\mathrm{total}} - M_{\mathrm{weight}} - M_{\mathrm{act}},\\
\sum_{b=1}^{BS} M_b \;\le\; M_{\mathrm{KValloc}},\qquad M_b = \kappa_{\mathrm{base}} \cdot |S_b|
\end{gathered}
\end{equation}

where \(M_\mathrm{total}\) is the total HBM capacity, \(M_\mathrm{weight}\) the weight footprint (determined only by the model and the parallelism scheme), \(M_\mathrm{act}\) the activation footprint, \(BS\) the batch-size, and \(|S_b|\) the sequence length of the \(b\)-th request. \(\kappa_\mathrm{base}\) is the per-token KV cache footprint under full residency (including indexer state):

\begin{equation}
\label{eq:kappabase}
\kappa_{\mathrm{base}} = \left(L_{\mathrm{KV}} \cdot D_{\mathrm{KV}} + L_{\mathrm{index}} \cdot D_{\mathrm{index}}\right) \cdot P
\end{equation}

where \(L_\mathrm{KV}\) and \(D_{KV}\) are the number of attention layers and the KV dimension, \(L_{index}\) and \(D_{index}\) the layer count and dimension of the DSA indexer (\(D_{index}=0\) without DSA), and \(P\) the number of bytes per element.

Even for GLM-5.2, which already adopts Multi-head Latent Attention (MLA) for dimensionality compression, the KV cache remains enormous. With BF16 (\(L_{KV}=78, D_{KV}=576, L_{index}=21, D_{index}=128\)),

\begin{equation}
\label{eq:kappabase-glm}
\kappa_\mathrm{base} = \left( 78 \times 576 + 21 \times 128 \right) \times 2 \mathrm{Bytes} = 95.23 \mathrm{kB},
\end{equation}

i.e., a per-request HBM footprint of \(M_b = 95.23 \mathrm{kB} \times |S_b|\): 6.09 GB, 48.76 GB, and 97.52 GB at the 64k, 512k, and 1M sequence lengths typical of agentic AI. Given typical per-GPU/NPU HBM capacities of 64--128 GB, the KV cache easily hits the capacity wall, posing distinct challenges to prefill and decode (Sections~\ref{sec:prefill-motivation} and \ref{sec:decode-motivation}).

\subsection{Motivation for Prefill KV Offloading}\label{sec:prefill-motivation}

During prefill, the KV cache of active requests grows as prompt chunks are processed. Keeping the complete history in HBM limits context length and concurrency while leaving less room for activations and new tokens. With \(H\) historical tokens and \(N\) newly scheduled tokens, full residency requires

\begin{equation}
\label{eq:prefill-constraint}
\kappa_{\mathrm{base}}(H+N)+M_{\mathrm{act}}(N)
\leq M_{\mathrm{avail}},
\end{equation}

Here, \(M_{\mathrm{avail}}\) is the HBM remaining after weights and fixed allocations. As \(H\) grows, the feasible \(N\) shrinks, forcing smaller chunks or fewer concurrent requests. More forward passes and scheduling rounds increase TTFT, while undersized chunks and batches reduce accelerator utilization and input throughput; sufficiently long requests cannot fit at all. For GLM-5.2-w4a8 on a two-server, 32-die Ascend 910C prefill node (DP2/TP16/EP32), the deployment reports a per-rank KV budget of 20.12~GiB after accounting for weights, activations, and runtime \cite{ref32}. A single 1M-token request (1,048,576 tokens) occupies about 93~GiB, exceeding this budget even at a batch size of one.

Unlike conventional prefix caching or swapping inactive requests, prefill therefore requires an offloading scheme that reduces the peak HBM footprint of live KV. Prefill is compute-intensive and exposes substantial per-layer computation that can overlap data movement. Live KV can be placed in host memory and loaded back into HBM one layer at a time when needed, while current-layer computation overlaps future-layer loading and completed-layer writeback. This directly lowers peak HBM and releases capacity for longer contexts, more concurrent requests, or larger chunks.

These capacity gains are useful only if KV movement does not introduce excessive exposed latency. Loading \(m_\ell\) bytes for layer \(\ell\) over achieved H2D bandwidth \(\beta_{\mathrm{H2D}}\) takes at least

\begin{equation}
\label{eq:prefill-h2d}
T_{\mathrm{H2D}}^{(\ell)}
\geq \frac{m_\ell}{\beta_{\mathrm{H2D}}}.
\end{equation}

Let \(T_{\mathrm{window}}^{(\ell)}\) be the independent computation between scheduling the preload and consuming its data. An approximate necessary condition for overlapping H2D with computation is

\begin{equation}
\label{eq:prefill-overlap}
T_{\mathrm{H2D}}^{(\ell)}+T_{\mathrm{sync}}^{(\ell)}
\lesssim T_{\mathrm{window}}^{(\ell)}.
\end{equation}

If H2D and D2H share a serialized copy path, the left-hand side must also include \(T_{\mathrm{D2H}}^{(\ell)}\). The challenge is to sustain this overlap with a small pool of device buffers: late preloads stall attention, while premature buffer reuse can overwrite KV still being consumed or written back. Layerwise offloading must therefore coordinate loading, computation, and writeback within both the HBM budget and the available computation windows; Section~\ref{sec:prefill} presents our design.

\subsection{Motivation of Decode Offloading}\label{sec:decode-motivation}

\begin{figure}[t]
\centering
\includegraphics[width=\columnwidth]{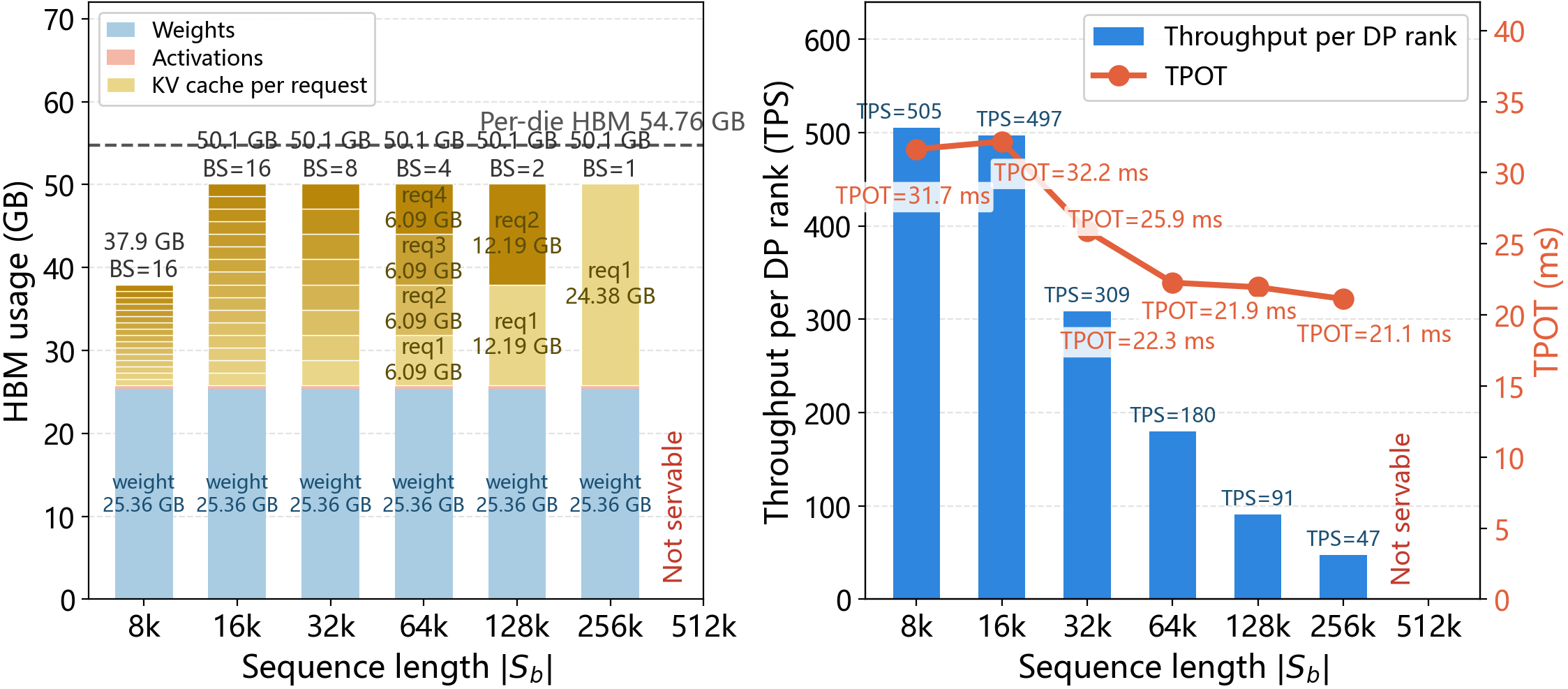}
\caption{The HBM capacity wall on decode nodes (dual-server 32-die Ascend910C, DP16/TP2/EP32, GLM-5.2-w4a8). Left: per-die HBM footprint breakdown; right: measured TPOT and throughput.}
\label{fig:mem-wall}
\end{figure}

The KV cache capacity problem makes decode nodes hit the HBM capacity wall: the batch-size, and hence throughput, is limited. Take the production deployment of GLM-5.2-w4a8 (dual-server 32-die Ascend910C, DP16/TP2/EP32): measured \(M_\mathrm{KValloc} = 29\) GB after deducting weights and activations, so a request of sequence length \(|S_b|\) allows a maximum decode batch-size per DP rank of \(BS_\mathrm{max} = \lfloor M_\mathrm{KValloc} / (\kappa_\mathrm{base} \cdot |S_b|) \rfloor\). The measurements in Figure~\ref{fig:mem-wall} show two consequences of the capacity wall: at sequence lengths \(\geq\)512k a single request no longer fits (ultra-long sequences cannot be served); at \(\geq\)32k, \(BS_\mathrm{max}\) drops to single digits, and since \(\mathrm{Throughput}_{\mathrm{decode}} = \mathrm{BS}/\mathrm{TPOT}\), throughput collapses (low long-sequence throughput).

The rise of DSA changes this premise. Since DeepSeek-V3.2 \cite{ref15} and GLM-5 \cite{ref16}, DSA has become mainstream: a lightweight LightningIndexer (LI) selects \(T_b\subset S_b\) with \(|T_b|=k\), and SparseFlashAttention (SFA) computes only over \(T_b\): \(T_b\leftarrow\mathrm{LI}(S_b)\), \(\mathrm{Attention\ Output}\leftarrow\mathrm{SFA}(T_b)\). With \(k=2048\) in GLM-5, computation falls to \(k/|S_b|\) of full attention.

Decode KV cache offloading with DSA can break the HBM capacity wall. Keep the full KV cache \(S_b\) in host DRAM and only a hot cache \(C_b\subset S_b\) in HBM. Each step loads \(\mathrm{Miss}_b=T_b-C_b\) while evicting equally many tokens, preserving lossless SFA over the fully resident \(T_b\). The per-request HBM footprint is \(M_b^\mathrm{offload}=(L_\mathrm{KV}D_\mathrm{KV}|C_b|+L_\mathrm{index}D_\mathrm{index}|S_b|)P\), or \(89.86\mathrm{kB}\times|C_b|+5.38\mathrm{kB}\times|S_b|\) for GLM-5.2. When \(|C_b|\ll|S_b|\), this is far below full residency.

A larger \(BS\), however, does not automatically increase throughput. Decode throughput is \(\mathrm{BS}/\mathrm{TPOT}=\mathrm{BS}\cdot N_\mathrm{spec}/\mathrm{ITL}\), where \(N_\mathrm{spec}\) is the average accepted tokens per step. Offloading gives \(\mathrm{ITL}_\mathrm{offload}=\mathrm{ITL}_\mathrm{baseline}+T_\mathrm{cache-manage}L_\mathrm{index}+T_\mathrm{H2D-load}L_\mathrm{KV}\), adding cache management (miss-lookup, pick eviction, etc.) and miss-loading latency. Baseline throughput already saturates at \(BS\approx32\), while \(T_\mathrm{H2D-load}\) grows with \(BS\). Suppressing these terms poses three challenges: (1) large miss count, requiring a larger \(|C_b|\) to reduce loads; (2) cache management latency that typically grows with \(|C_b|\), opposing that remedy; and (3) H2D loads serialized between LI and SFA, since loading awaits top-\(k\) selection and SFA awaits loading.

\subsection{Related Work}\label{sec:related}

\textbf{Prefill offloading.} FlexGen jointly places weights, activations, and KV state across GPU, CPU, and storage to execute models under constrained device-memory budgets \cite{ref2}. D\'ej\`aVu streams KV state for prefill--decode disaggregation and fault tolerance \cite{ref3}; Mooncake \cite{ref25} and LMCache \cite{ref21} organize reusable KV cache around disaggregated storage. These systems expand effective cache capacity or migrate state between serving components, but target reusable or inactive KV rather than reducing the peak HBM of active KV while a request executes prefill --- precisely the capacity constraint analyzed in Section~\ref{sec:prefill-motivation}.

\textbf{Decode offloading.} Closest to this work are two offloading systems for natively sparse models, HiSparse \cite{ref30} and ECHO \cite{ref20}. Both share a structural limitation in addressing the low-latency challenge: cache management overhead grows with the cache budget (or the top-\(k\) size), is paid at every layer of every step, and cannot be decoupled from cache capacity. HiSparse checks hits via a shared-memory hash intersection (build \(O(k)\) + full scan \(O(|C_b|)\)) and evicts by rewriting the entire LRU order each step (\(O(|C_b|)\)), with measured miss rates as high as 13\%--30\%. ECHO performs an independent miss-lookup after top-\(k\) (\(O(BS \cdot k)\) per step per layer), evicts via a radix selection over the whole cache pool (\(O(|C_b|)\)), and relies on EMA prediction of the top-\(k\) scores for load overlap. AVSG \cite{ref35} follows the same lossless DSA-offloading approach, preserving exact token selections while reusing an HBM buffer across decode steps to reduce H2D traffic. ESS \cite{ref33} first proposed DualBatch-Attention (DBA) overlap, splitting LI along the batch dimension and SFA into hit and miss parts to create windows for H2D loads. However, ESS overlooks cache management overhead, and splitting the kernels incurs additional latency. Our design instead folds the miss-lookup into the top-\(k\) process itself (zero extra miss-lookup overhead) and overlaps loads with exact (non-predictive) intra-kernel pipelining. Other sparse offloading systems mainly target non-natively-sparse models \cite{ref26} and reduce traffic through prediction, auxiliary representations, or vector retrieval \cite{ref24}, including InfiniGen, ShadowKV, FreeKV, and RetrievalAttention \cite{ref4,ref11,ref31,ref12}, which also alter the model-defined attention set.

\section{Phase-Specialized Unified KV Offloading Architecture}\label{sec:architecture}

We propose a phase-specialized yet unified KV offloading architecture. Prefill and decode share one logical KV history backed by host DRAM, together with paged host blocks, block tables, and a common control plane, while specializing device residency and transfer granularity. Committed KV resides in host memory, and the device retains only the bounded working set required by the active phase. At the phase boundary, the host-side KV history is reused or transferred according to the deployment, while device residency switches from full-layer buffers to a token-level hot cache.

\begin{figure}[t]
\centering
\includegraphics[width=0.45 \textwidth]{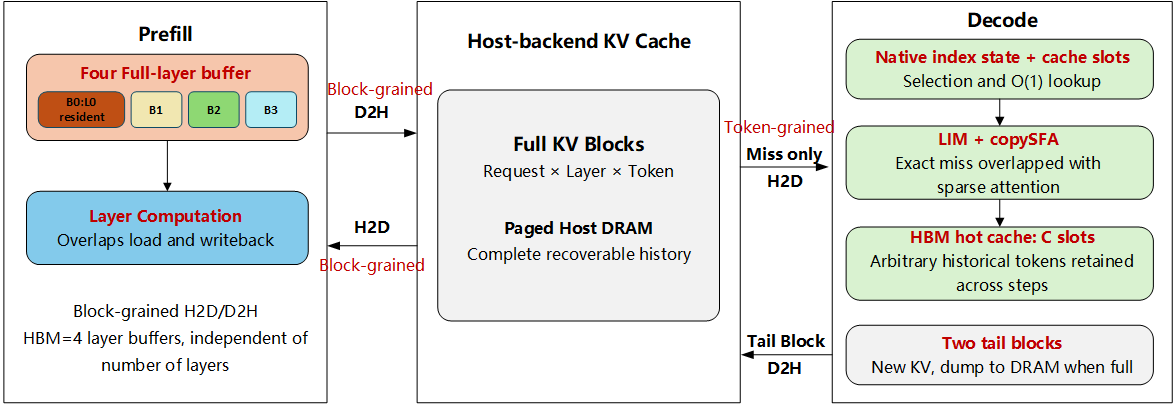}
\caption{Phase-specialized unified KV offloading architecture.}
\label{fig:architecture}
\end{figure}

Prefill uses four full-layer buffers: \(B_0\) remains resident for the first layer, while \(B_1\)--\(B_3\) rotate across later layers to pipeline block-grained H2D loads, layer computation, and D2H writebacks \cite{ref17,ref18}. Decode instead retains the native indexer, a \(C\)-slot hot cache, two new-KV tail blocks, and the \texttt{cache\_slots} residency map in HBM. LIM and CopySFA coordinate selection, exact miss retrieval, and sparse attention so that only model-selected misses are transferred. Only decode depends on a native sparse indexer; layerwise prefill loads and writes back complete layer KV and therefore also applies to dense attention. Figure~\ref{fig:architecture} shows the two phase-specialized paths and their shared host-backed state.

Let \(s_{\mathrm{KV}}\) be the per-token, per-layer K/V bytes on one tensor-parallel rank, \(T\) the context length, \(L\) the layer count, \(C\) the per-layer decode hot-cache capacity, and \(b\) the page size. A device-resident baseline requires

\begin{equation}
\label{eq:mbase}
M_{\mathrm{base}} = LTs_{\mathrm{KV}}.
\end{equation}

The dominant phase-specific allocations are approximately

\begin{equation}
\label{eq:mstages}
\begin{aligned}
M_{\mathrm{prefill}} &\approx 4Ts_{\mathrm{KV}},\\
M_{\mathrm{decode}} &\approx L(C+2b)s_{\mathrm{KV}}+M_{\mathrm{index}}(T)\\
&\quad +M_{\mathrm{slots}}(T)+M_{\mathrm{meta}}.
\end{aligned}
\end{equation}

Thus, prefill retains a dominant K/V fraction of \(4/L\), or about 5.1\% for \(L=78\). Decode retains approximately \((C+2b)/T\); with \(C/T=1/4\) and \(b\ll T\), this approaches a fourfold reduction. Because the native index, \texttt{cache\_slots}, and other metadata may still scale with \(T\), these expressions quantify the dominant K/V-data savings and slightly overestimate total HBM savings.

\section{Layerwise Prefill KV Cache Offloading}\label{sec:prefill}

This section presents the four-buffer design, followed by its per-layer lifecycle, asynchronous schedule, and overlap condition.

\subsection{Design Overview}\label{sec:prefill-logical}

The design exploits Transformer's layerwise lifetime: although every layer owns KV state, only the current layer consumes its complete KV tensor at a time. The host store preserves the full history by request, layer, and paged token block, while the device allocates only four full-layer buffers \cite{ref17,ref18}. \(B_0\) remains bound to the first layer \(L_0\) throughout prefill, and a binding table rotates \(B_1\)--\(B_3\) across all later layers. Indexer state and block-table metadata are stored separately and are not overwritten as the buffers rotate.

Keeping the first layer resident removes the H2D wait for \(L_0\) and uses its computation to launch the subsequent load--compute--writeback pipeline. Later layers retain bandwidth-efficient block transfers, so greater model depth enlarges host storage without increasing the number of device buffers.

\subsection{Per-Layer Execution Lifecycle}\label{sec:prefill-lifecycle}

The implementation reuses the layer hooks exposed by vLLM KVConnector. \(L_0\) directly uses resident \(B_0\); offloaded layers \(\ell\geq1\) follow this lifecycle through \texttt{wait\_for\_layer\_load} and \texttt{save\_kv\_layer}:

\begin{enumerate}
\def\labelenumi{\arabic{enumi}.}
\tightlist
\item
  \textbf{Bind and load.} The connector selects a free buffer from \(B_1\)--\(B_3\), binds it to layer \(\ell\), and asynchronously loads valid history according to the request block tables. A new request may skip the load, while chunked prefill loads only allocated blocks preceding the current chunk.
\item
  \textbf{Wait for readiness.} Before attention reads history, \texttt{wait\_for\_layer\_load} waits for that buffer's H2D event.
\item
  \textbf{Append and compute.} New K/V for the current chunk is written directly to scheduler-provided slots in the bound buffer. The original prefill attention then consumes both history and appended entries without another staging tensor.
\item
  \textbf{Write back and confirm completion.} After the current layer finishes, \texttt{save\_kv\_layer} asynchronously writes the updated KV to host memory through block-grained D2H. The host KV cannot be consumed by a later stage until D2H completes.
\item
  \textbf{Release and prefetch.} A buffer becomes reusable only after its last compute consumer and D2H writeback complete; the completion event may also launch the future load selected by \texttt{onload\_layer\_map}.
\end{enumerate}

This lifecycle applies to every active prefill batch. Continuous batching may update request block tables, but event dependencies preserve buffer ownership. Before decode begins, all writebacks and index-state updates must complete.

\subsection{Asynchronous Pipeline Schedule and Overlap Conditions}\label{sec:prefill-pipeline}

When the platform provides independent engines, the runtime uses separate compute, H2D, and D2H streams. Compute starts immediately with \(L_0\) in resident \(B_0\) while \(L_1\) loads into \(B_1\). In steady state, compute executes the current layer, H2D preloads a future layer, and D2H commits a completed layer; the final iterations drain outstanding writebacks.

\texttt{onload\_layer\_map[i]=j} specifies the future layer \(j\) to preload when layer \(i\) reaches its scheduling point. It combines the three rotating buffers with a chosen prefetch distance and need not select \(i+1\). A slower H2D may start several layers ahead, but the target buffer must have completed computation and writeback. Device events make compute wait for H2D readiness and reuse wait for compute and D2H completion; if no buffer is free, the pipeline waits rather than overwriting live state.

\begin{figure}[t]
\centering
\includegraphics[width=0.8 \columnwidth]{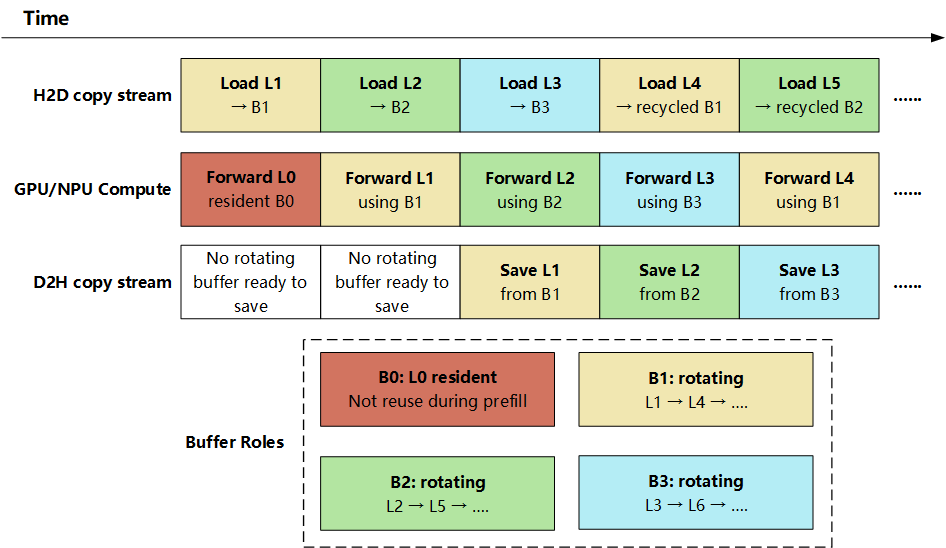}
\caption{Four-buffer prefill schedule. The timeline is schematic rather than a measured trace.}
\label{fig:prefill-pipeline}
\end{figure}

Figure~\ref{fig:prefill-pipeline} illustrates the schedule: \(L_0\) starts on \(B_0\), after which \(B_1\)--\(B_3\) rotate so that loading, computation, and writeback progress concurrently.

The overlap achieved by this pipeline depends on the compute windows available for prefetching and buffer reuse. Let \(d_{\mathrm{on},\ell}\) be the independent layer computation between preloading layer \(\ell\) and consuming it, and \(d_{\mathrm{off},\ell}\) the computation available before its buffer is reused. With independent H2D and D2H engines, exposed latency satisfies

\begingroup
\setlength{\jot}{1pt}
\begin{gather}
	\tau_{\mathrm{on,exposed}}^{(\ell)} \geq
	\max\!\left(0,\tau_{\mathrm{H2D}}^{(\ell)}-
	d_{\mathrm{on},\ell}\tau_{\mathrm{compute}}\right) \label{eq:tauon} \\
	\tau_{\mathrm{off,exposed}}^{(\ell)} \geq
	\max\!\left(0,\tau_{\mathrm{D2H}}^{(\ell)}-
	d_{\mathrm{off},\ell}\tau_{\mathrm{compute}}\right) \label{eq:tauoff}
\end{gather}
\endgroup

If both directions share a serialized copy path, the conservative condition uses \(\tau_{\mathrm{move}}=\tau_{\mathrm{H2D}}+\tau_{\mathrm{D2H}}\). When movement exceeds the available window, \texttt{onload\_layer\_map} can increase prefetch distance, or more layers can remain resident \cite{ref18}. Reserving \(B_0\) sacrifices one rotating slot but removes the first-layer load and immediately creates the computation window that starts the pipeline.

\section{Low-Latency Decode KV Cache Offloading}\label{sec:decode}

This section expands the decode part of the unified KV cache offloading architecture of Section~\ref{sec:architecture}: Section~\ref{sec:decode-overall} presents the overall design, and Sections~\ref{sec:cache-slots}--\ref{sec:transfer-overlap} address the three latency challenges raised in Section~\ref{sec:decode-motivation}.

\subsection{Overall Design}\label{sec:decode-overall}

\begin{figure}[t]
\centering
\includegraphics[width=1.0 \columnwidth]{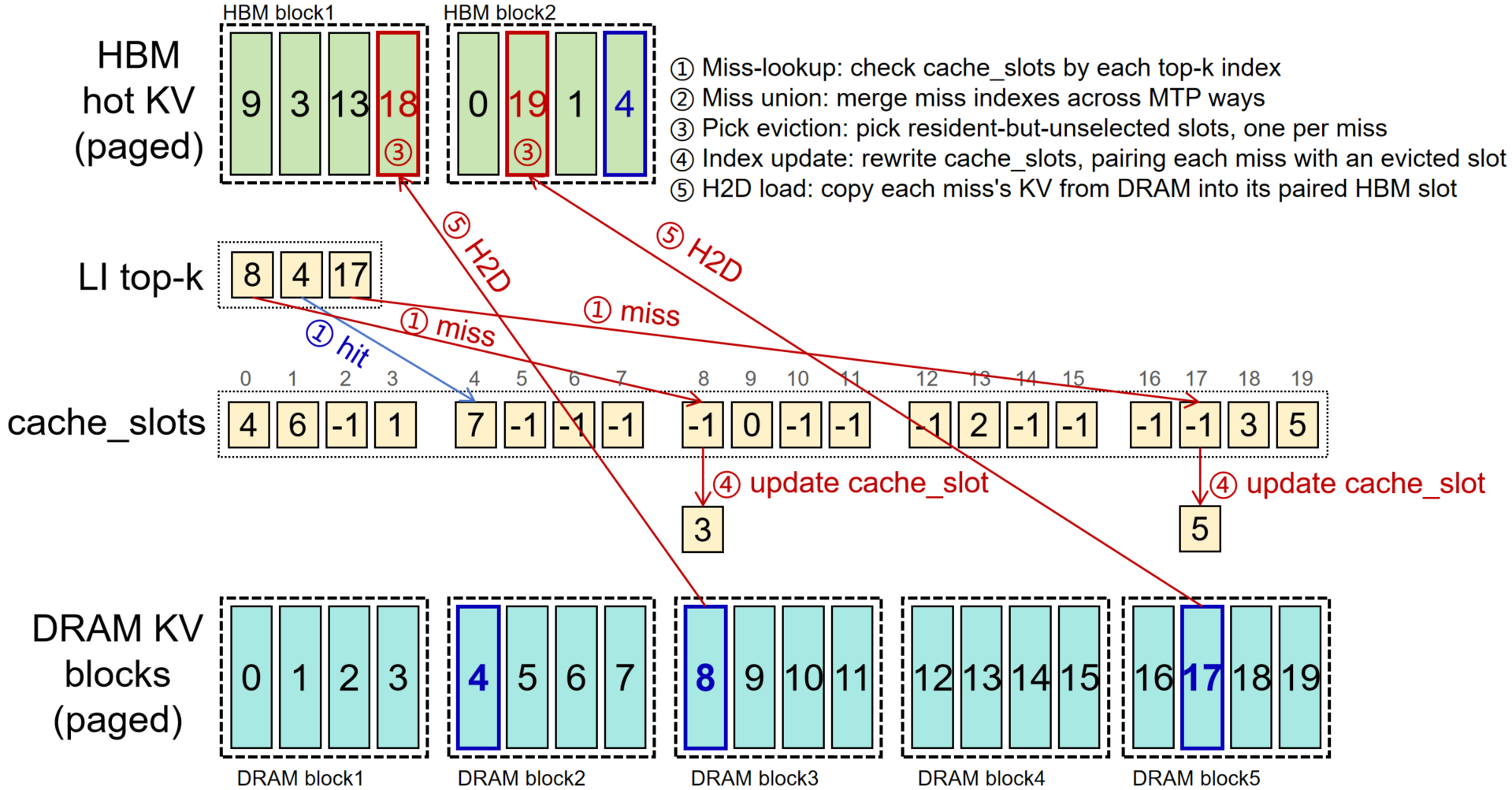}
\caption{Cache data layout and cache management flow of one steady-state decode step (single-request, single-way top-\(k\) view)}
\label{fig:cache-slots}
\end{figure}

We modify vLLM-Ascend so that decode nodes execute the decode KV cache offloading of Section~\ref{sec:decode-motivation} \cite{ref19}. As shown in Figure~\ref{fig:cache-slots}, the full KV cache \(S_b\) of request \(b\) is kept in host DRAM, while HBM holds only the hot cache \(C_b \subset S_b\). The design keeps vLLM's block management: DRAM blocks hold all tokens in their original order, while HBM blocks hold scattered hot tokens (\(C_b\)).

From the decode node's perspective, a request's lifecycle is: after prefill finishes, its KV cache is written into the decode node's host DRAM KV blocks (P\(\rightarrow\)D transfer); the first decode step loads \(|C_b|\) tokens from DRAM into the HBM KV blocks (first-step fill); subsequent steps enter steady state, executing the cache swap-in and eviction flow at every layer of every step. Under MTP3, the steady-state decode flow is:

\begin{equation}
\label{eq:steady-flow}
\resizebox{\columnwidth}{!}{$
\begin{aligned}
T_b^{(j)}               & \leftarrow \mathrm{LI}(S_b), ~~ j = 0 \sim 3 ~ (\mathrm{MTP3})   &  & \text{where} ~~ T_b^{(j)} \subset S_b \\
\mathrm{Miss}_b^{(j)}   & \leftarrow T_b^{(j)} - C_b                                       &  & \text{\stepnum{1}~Miss-lookup: HBM residency per way} \\
\mathrm{Miss}_b         & \leftarrow \bigcup\nolimits_j \mathrm{Miss}_b^{(j)}              &  & \text{\stepnum{2}~Miss union: merge 4-way misses} \\
\mathrm{Evict}_b        & \leftarrow \mathrm{Evict}\left( C_b - \bigcup\nolimits_j T_b^{(j)} \right) &  & \text{\stepnum{3}~Pick eviction}, ~~ |\mathrm{Evict}_b| = |\mathrm{Miss}_b| \\
C_b^\mathrm{new}        & \leftarrow C_b - \mathrm{Evict}_b \cup \mathrm{Miss}_b           &  & \text{\stepnum{4}~Index update: pair miss to evicted slot} \\
\mathrm{Miss}_b         & \xrightarrow{\mathrm{H2D}} \mathrm{Evict}_b ~ \text{slots}          &  & \text{\stepnum{5}~H2D load: DRAM to HBM per pairing} \\
\mathrm{Output}_b^{(j)} & \leftarrow \mathrm{SFA}(T_b^{(j)})                               &  & \text{where} ~~ T_b^{(j)} \subset C_b^\mathrm{new}
\end{aligned}$}
\end{equation}

where \(\mathrm{Miss}_b\) is the deduplicated union of the 4-way misses (i.e., this step's load volume) and \(\mathrm{Evict}_b\) is paired with it one-to-one, so the hot-cache size is unchanged (\(|C_b^\mathrm{new}| = |C_b|\)); \stepnum{1}--\stepnum{5} are the five steps of this flow, referenced by these numbers in the following subsections.

Figure~\ref{fig:cache-slots} illustrates this flow with a single request and a single-way top-\(k=\{8,4,17\}\) (marking each step's location with token17). To record which tokens reside in HBM, we design the cache index \textbf{cache\_slots}: a 1-D int32 tensor of length \(|S_b|\) per request per layer recording each token's resident slot (-1 if not resident).

\begin{figure}[t]
\centering
\includegraphics[width=0.7 \columnwidth]{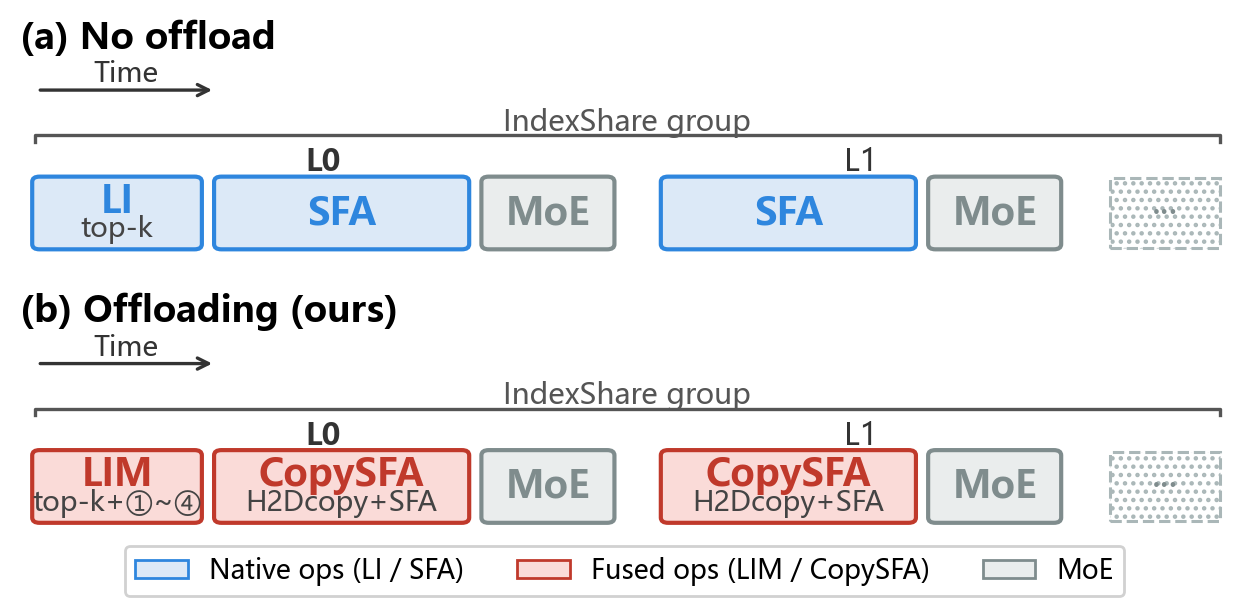}
\caption{Kernel pipelines of one steady-state decode step in GLM-5.2.}
\label{fig:decode-pipeline}
\end{figure}

FlashOffloading's kernel flow (Figure~\ref{fig:decode-pipeline}(b)) differs from the non-offloaded flow (Figure~\ref{fig:decode-pipeline}(a)) in two changes --- the LI kernel is replaced by the fused \textbf{LIM} kernel (which performs the cache management of steps \stepnum{1}--\stepnum{4} while computing top-\(k\)), and the SFA kernel is replaced by the fused \textbf{CopySFA} kernel (which performs the step-\stepnum{5} H2D load while executing SFA).

\subsection{Cache Indexing: cache\_slots}\label{sec:cache-slots}

Enlarging \(|C_b|\) reduces misses, but prior slot\(\rightarrow\)token indices require step \stepnum{1} to scan \(O(|C_b|)\) entries, yielding \(T_\mathrm{cache-manage}\sim O(k)+O(|C_b|)\). A small cache therefore increases H2D traffic, while a large one increases management latency.

Our reverse token\(\rightarrow\)slot \textbf{cache\_slots} index makes each miss-lookup an \(O(1)\) lookup, reducing management to \(T_\mathrm{cache-manage}\sim O(k)+O(|\mathrm{Miss}_b|)\), where \(|\mathrm{Miss}_b|\ll|C_b|\). Pick eviction and index updates also touch only misses, decoupling latency from capacity: \(|C_b|/|S_b|\) can grow from \(1/8\) to \(1/4\), supporting nearly \(4\times\) larger batches (Section~\ref{sec:cache-strategy}).

\subsection{Low-Latency Cache Management}\label{sec:cache-management}

\begin{figure}[t]
\centering
\includegraphics[width=0.8 \columnwidth]{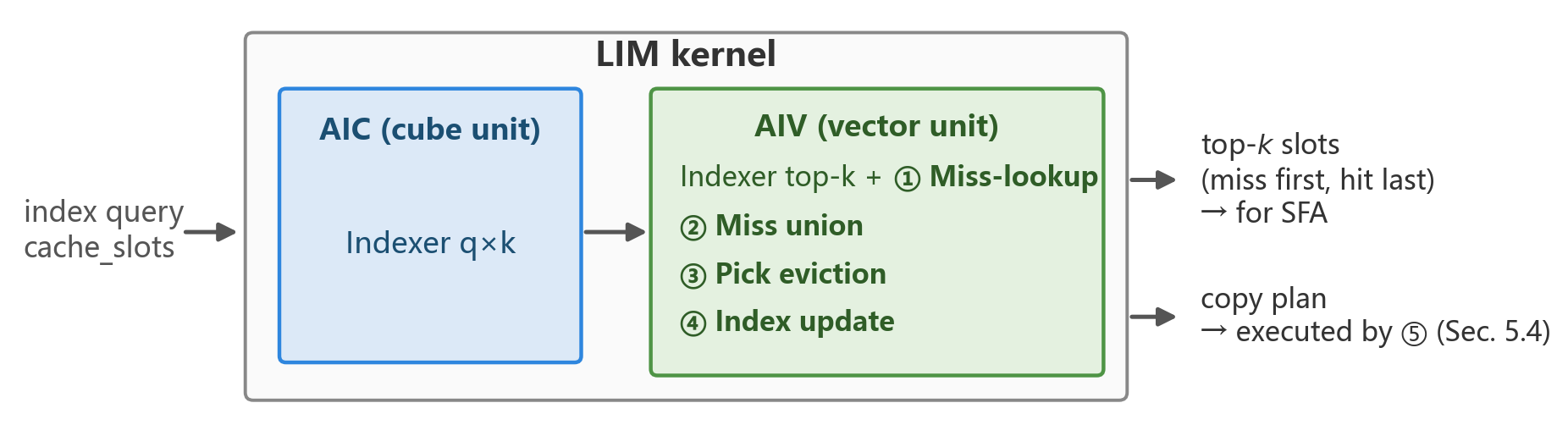}
\caption{Internal data flow of the LIM kernel}
\label{fig:lim-flow}
\end{figure}

Section~\ref{sec:cache-slots} compressed \(T_\mathrm{cache-manage}\) to \(O(k)+O(|\mathrm{Miss}_b|)\) at the index-structure level; this section further compresses the miss-lookup step (the \(O(k)\) term). The conventional approach waits for top-\(k\) to finish, then performs a cache\_slots miss-lookup per selected index --- \(k\) extra memory accesses per way. Our approach encodes each candidate token's index and its cache\_slots value into a single int32 before top-\(k\), which participates in the selection together with the scores; after selection, index and slot are decoded from the winning entries, and the miss result follows naturally from top-\(k\). Taking the 20 candidate tokens of Figure~\ref{fig:cache-slots} as an example:

{\fontsize{4.5}{5.5}\selectfont
\begin{verbatim}
index       = [0, 1, 2, 3, 4, 5, 6, 7, 8, 9, 10, 11, 12, 13, 14, 15, 16, 17, 18, 19]
cache_slots = [4, 6, -1, 1, 7, -1, -1, -1, -1, 0, -1, -1, -1, 2, -1, -1, -1, -1, 3, 5]
# before top-k: encode index and cache_slots pairwise; after: decode slot and index
topk_index  = [8, 4, 17]    # i.e., T_b
topk_slot   = [-1, 7, -1]
\end{verbatim}
}

topk\_slot directly gives the miss-lookup result (token8 and token17 miss; token4 hits slot 7), with miss-lookup overhead eliminated by fusion into LI's top-\(k\). This optimization is implemented as the Ascend kernel \textbf{LIM} (LightningIndexerManage, Figure~\ref{fig:lim-flow}), whose outputs feed the downstream CopySFA kernel along two paths: the top-\(k\) slot list for the attention computation, and copy triples (DRAM source, HBM destination, count) for the step-\stepnum{5} H2D load.

\subsection{Transfer Overlap}\label{sec:transfer-overlap}

After cache management completes, \(\mathrm{Miss}_b\) must be loaded from DRAM to HBM (step \stepnum{5}). Prior work ESS \cite{ref33} splits SFA coarsely into a hit-then-miss pair of kernels to overlap the load, but the split breaks the intra-kernel pipeline and overlaps poorly; we instead fuse the H2D load and SFA into a single CopySFA kernel, overlapping at fine granularity inside the kernel.

\begin{figure}[t]
\centering
\includegraphics[width=0.85 \columnwidth]{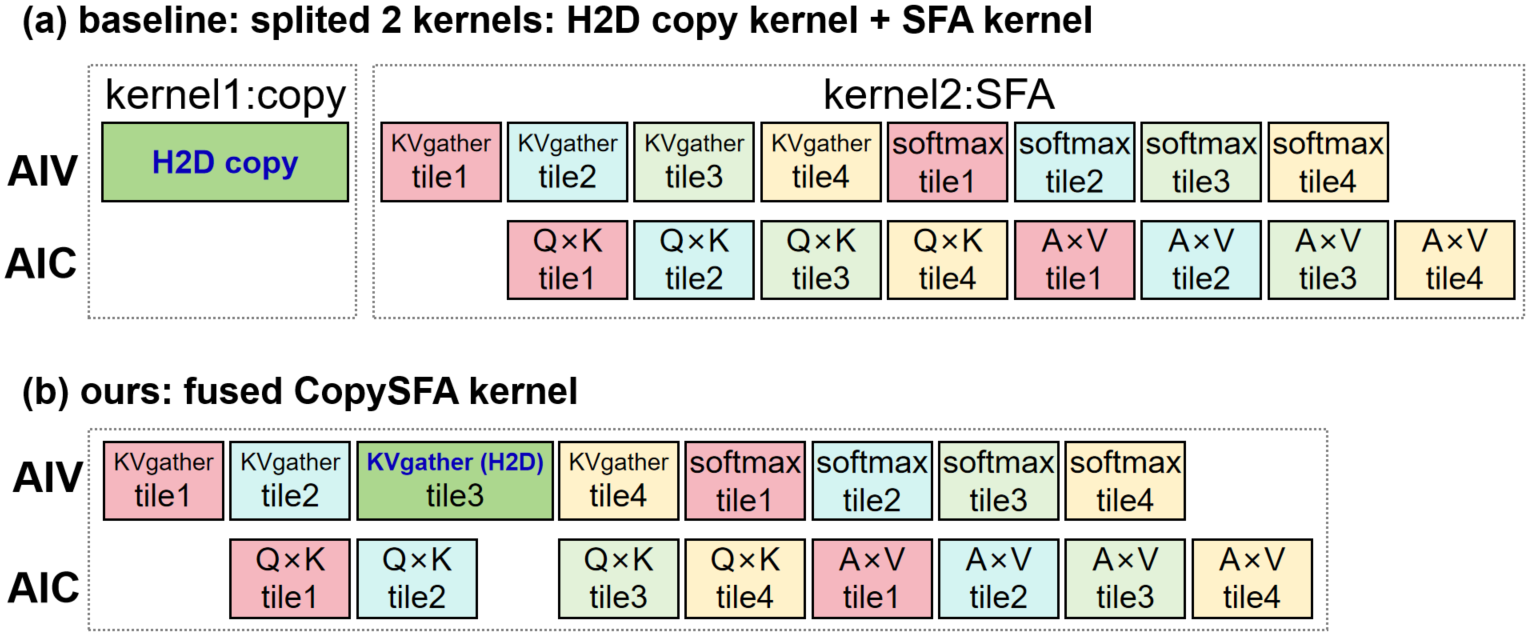}
\caption{Intra-kernel pipelines of the non-overlapped and overlapped schemes}
\label{fig:copysfa-pipeline}
\end{figure}

CopySFA's key lies in SFA's KVgather step (Figure~\ref{fig:copysfa-pipeline}(a)): SFA splits the 2048 tokens into 4 tiles, each executing KVgather (moving the scattered HBM tokens to a contiguous workspace), Q\(\times\)K, softmax, and A\(\times\)V in turn. We branch on miss/hit inside KVgather: miss tokens are fetched from host pin-memory over PCIe/HCCS to on-chip, while hit tokens move from HBM as before (Figure~\ref{fig:copysfa-pipeline}(b)); and since LIM has gathered the miss tokens at the head, only 1 of the 4 tiles involves DRAM movement.

The fused kernel (Figure~\ref{fig:copysfa-pipeline}(b)) has lower latency than the separate scheme (Figure~\ref{fig:copysfa-pipeline}(a)) for two reasons: (1) one fewer HBM read pass --- the separate scheme's miss tokens go through ``read DRAM \(\rightarrow\) write slots \(\rightarrow\) read slots \(\rightarrow\) write workspace'' (4 passes in total), while CopySFA directly ``reads DRAM \(\rightarrow\) writes both slots and workspace'' (3 passes); (2) the H2D load is overlapped with AIC computation --- in the separate scheme the AIC idles during the H2D load, while in CopySFA the DRAM movement runs in parallel with the hit tiles' Q\(\times\)K.

\section{Evaluation}\label{sec:evaluation}

This section validates FlashOffloading: Section~\ref{sec:setup} describes the experimental setup; Section~\ref{sec:prefill-ablation} shows that larger chunk prefill sizes improve prefill throughput; Section~\ref{sec:decode-ablation} ablates decode offloading to independently validate its three optimizations; and Section~\ref{sec:e2e} evaluates end-to-end inference in a real PD-disaggregated scenario.

\subsection{Experimental Setup}\label{sec:setup}

\textbf{Platform.} All experiments run on Ascend910C NPUs (2 dies per card, 64GB HBM per die; see Section~\ref{sec:prefill-motivation}). The kernel-level unit tests (Sections~\ref{sec:index-latency}, \ref{sec:overlap-effect}) run on a single die; the end-to-end experiments (Section~\ref{sec:e2e}) run on a four-server 32-card 64-die cluster with 1P1D PD disaggregation: prefill node 32 dies (DP2/TP16/EP32), decode node 32 dies (DP16/TP2/EP32), and after the P node finishes prefill, the KV cache is written to the D node's host DRAM via H2H transfer.

\textbf{Model and framework.} The model is GLM-5.2-w4a8c16, with MTP3 uniformly enabled for decode; the inference framework is a modified vLLM 0.30 + vLLM-Ascend 0.30, on which prefill/decode offloading can be toggled independently.

\textbf{Decode steady-state operating point.} The kernel unit tests of Section~\ref{sec:decode-ablation} share a typical steady-state decode operating point: HBM hot-cache budget \(|C_b|=12288\) tokens and a per-request miss union of about 300 tokens (from the trace measurements of Section~\ref{sec:cache-strategy}); the HBM cache budget of Section~\ref{sec:e2e} ranges over 8192--30720 tokens depending on batch-size.

\subsection{Prefill Offloading Ablation}\label{sec:prefill-ablation}

\textbf{Method.} The platform follows Section~\ref{sec:setup}. The client sends one request with 256000 input tokens and one output token (prefill only), with a 50\% prefix hit rate: the first 128000 tokens reuse the prefix cache in DRAM, and the remaining 128000 are recomputed. We vary max-num-batched-tokens (i.e., chunk prefill size) and compare P90 TTFT with prefill offloading disabled and enabled.

\begin{figure}[t]
\centering
\includegraphics[width= 0.85 \columnwidth]{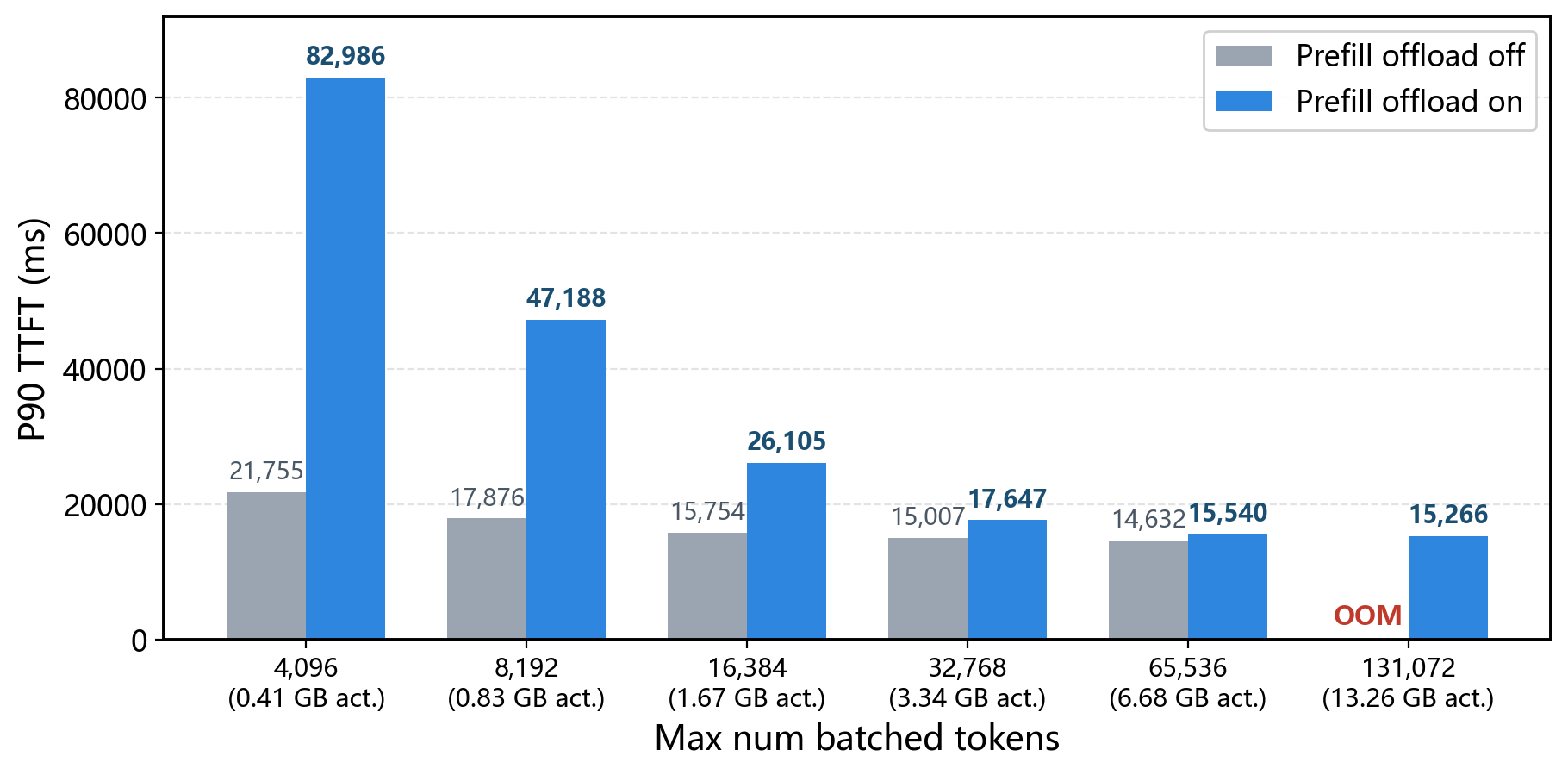}
\caption{P90 TTFT across chunk prefill sizes.}
\label{fig:prefill-chunk-ttft}
\end{figure}

Figure~\ref{fig:prefill-chunk-ttft} yields three observations:

\begin{itemize}
\tightlist
\item
  Larger max-num-batched-tokens lowers TTFT through higher compute utilization.
\item
  At the same max-num-batched-tokens, prefill offloading increases TTFT: without offloading, the reused prefix is loaded only for the first model.forward; with offloading, it is loaded in every forward pass.
\item
  Prefill offloading reduces KV cache HBM usage, leaving more room for activations and thus larger max-num-batched-tokens. Consequently, offloading with a larger setting can outperform no offloading with a smaller one.
\end{itemize}

\subsection{Decode Offloading Ablation}\label{sec:decode-ablation}

This section performs policy-level and kernel-level ablations of the three decode-side optimizations, validating that each contributes independently.

\subsubsection{Cache Strategy and Miss Count}\label{sec:cache-strategy}

This section validates the claim of Section~\ref{sec:cache-slots}: enlarging the cache budget \(|C_b|\) reduces the miss count more than sophisticated eviction policies (e.g., HiSparse's LRU).

\textbf{Method.} We run GLM-5.2 with MTP3 on Open-SWE long-context tasks and collect the decode-time top-2048 indices \(T_b^{(j)}\) (\(j=0\sim3\)) of every layer; an offline simulator then replays the steady-state swap-in/swap-out flow of Section~\ref{sec:decode-overall} on the same access sequence, varying only the eviction policy of step \stepnum{3}: Random (random pick), LRU (least-recently selected), Score (evict low-score tokens), and LIM (ours) (score threshold + chunk scan). The dataset is 20 Open-SWE requests: 10 short-sequence (66K--92K tokens) and 10 long-sequence (102K--131K tokens).

\begin{figure}[t]
\centering
\includegraphics[width=0.9 \columnwidth]{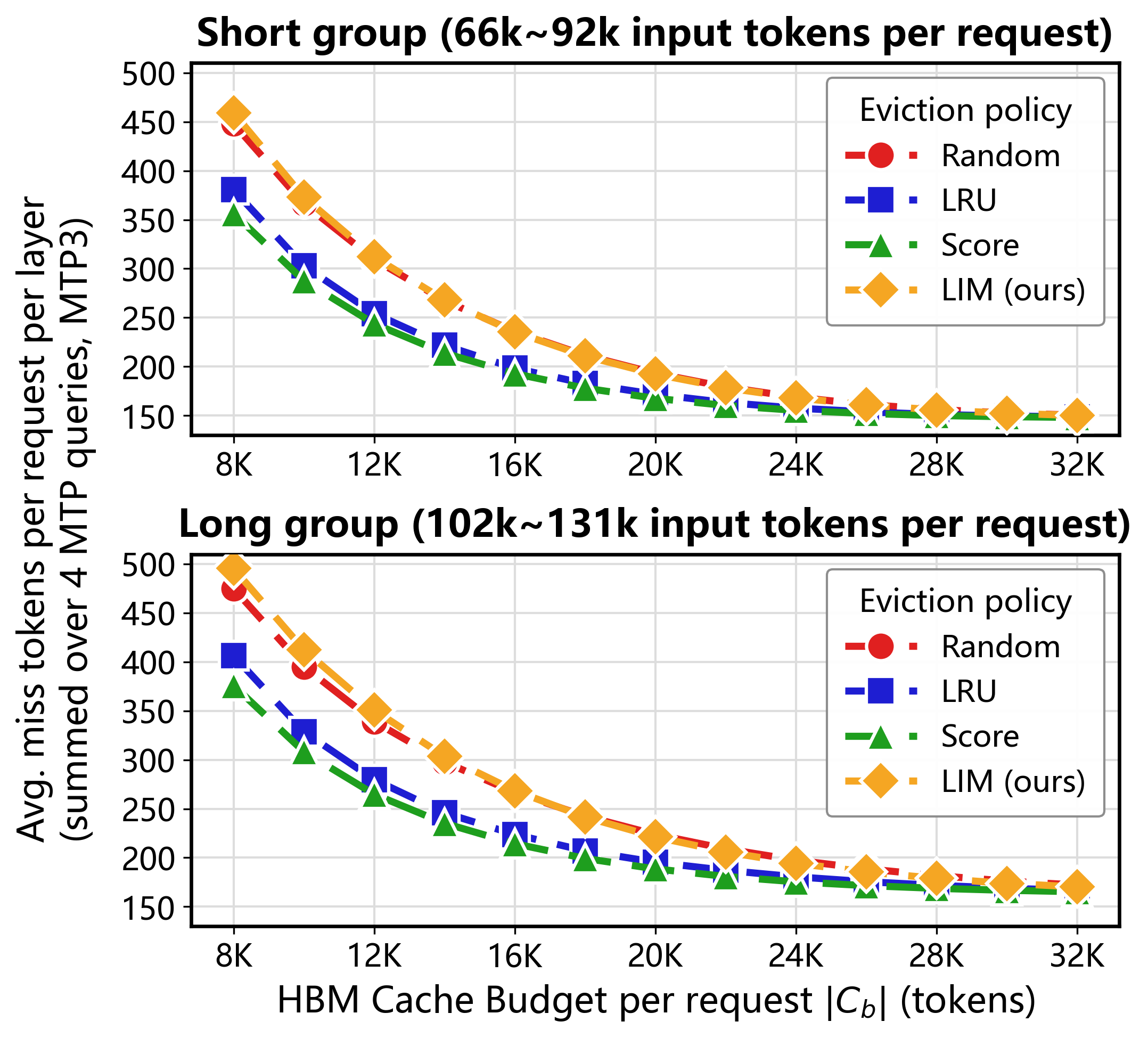}
\caption{Average miss tokens per request per layer vs.~cache budget and eviction policy}
\label{fig:cache-strategy}
\end{figure}

The results (Figure~\ref{fig:cache-strategy}) show that cache capacity is the primary determinant of the miss count, and the eviction policy a secondary one: enlarging \(|C_b|\) from 8192 to 16384 reduces misses by 43\%--49\%, while at \(|C_b|=16384\) Score reduces misses by only another 18\%--20\% relative to LIM, and the policy gap converges as capacity grows. Our typical configuration (LIM, \(|C_b|=14336\)) has 268--350 misses, below HiSparse's typical configuration (LRU, \(|C_b|=8192\)) at 380--406; although a larger \(|C_b|\) occupies more HBM, it stays under 1/4 of the sequence length, enough to support a \(\sim\)4\(\times\) batch-size increase --- and the measurements in Section~\ref{sec:decode-motivation} show that throughput already saturates at batch-size\(\approx\)32. This work turns precisely the HBM that prior work could not exploit into a lower miss count.

\subsubsection{Index Management Latency}\label{sec:index-latency}

This section validates the low latency of the Section~\ref{sec:cache-management} cache management scheme.

\textbf{Method.} On a single die (Section~\ref{sec:setup}), we unit-test native LI and LIM: both execute the same score computation and top-\(k\), while LIM additionally performs the cache management of \stepnum{1}--\stepnum{4}; the latency difference is the management overhead. The operating point follows Section~\ref{sec:setup}.

\begin{figure}[t]
\centering
\includegraphics[width=0.72 \columnwidth]{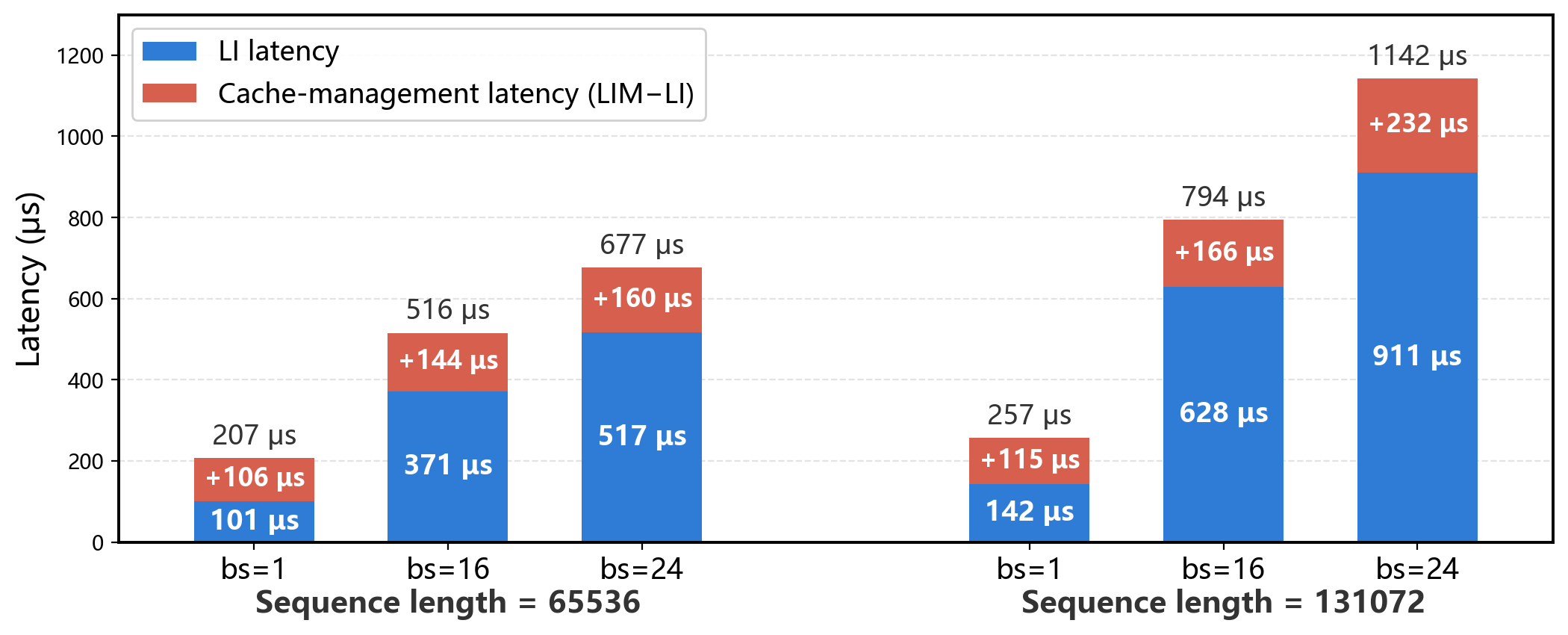}
\caption{LIM vs.~native LI kernel latency (single Ascend910C die, MTP3, cache budget 12288 tokens, per-request miss union 300 tokens).}
\label{fig:lim-latency}
\end{figure}

The results (Figure~\ref{fig:lim-latency}) show the extra cache management latency is only on the order of 106--232\ensuremath{\mu}s. End to end, only 21 of GLM-5.2's 78 layers have an LI/LIM kernel, so each decode step adds only about 160\ensuremath{\mu}s\(\times\)21\(\approx\)3.4 ms --- a small fraction of the \(\sim\)100 ms total ITL.

\subsubsection{H2D Load Overlap Effect}\label{sec:overlap-effect}

\begin{figure}[t]
\centering
\includegraphics[width=0.75 \columnwidth]{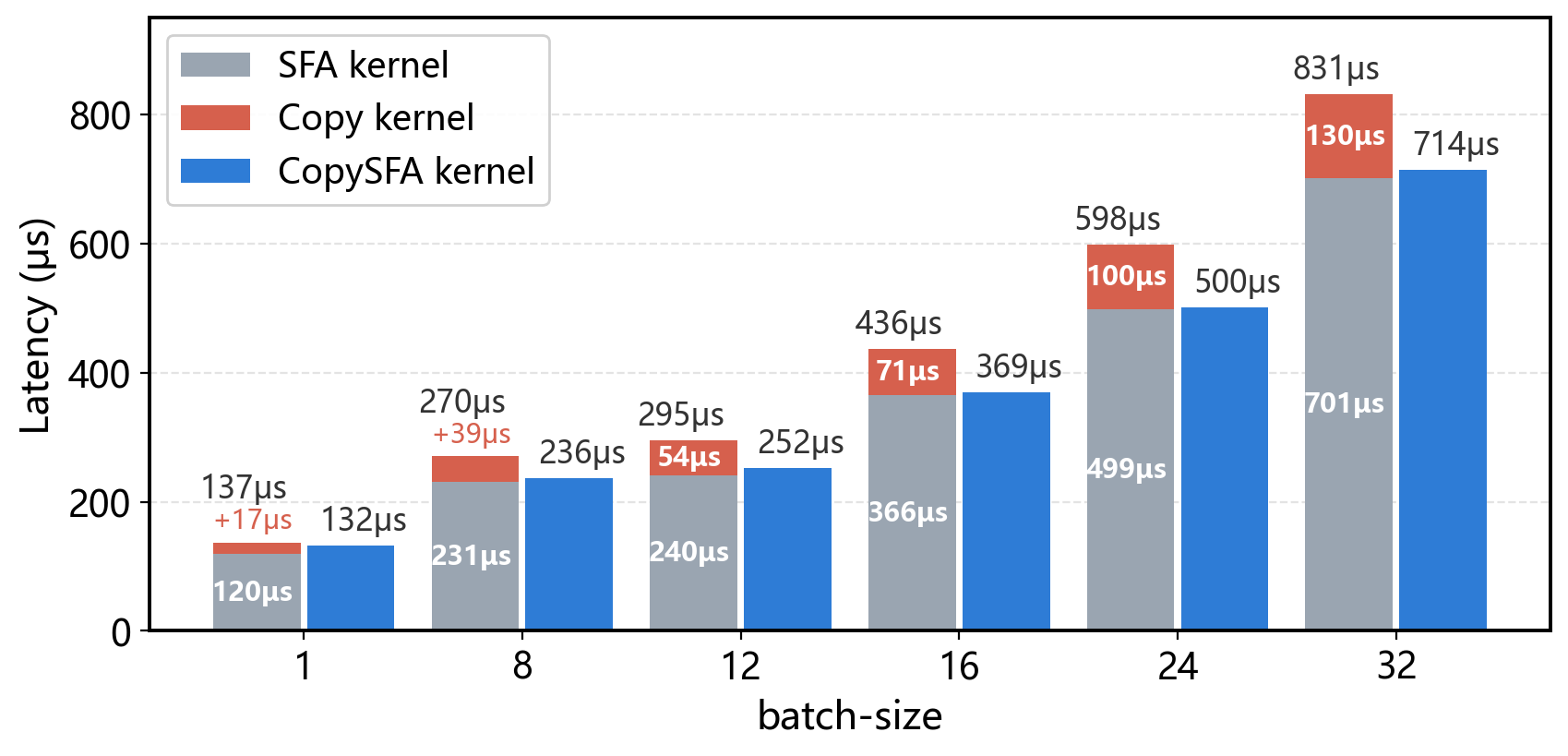}
\caption{CopySFA overlapped kernel vs.~non-overlapped baseline (single Ascend910C die, MTP3, per-request miss union 300 tokens).}
\label{fig:copysfa-latency}
\end{figure}

This section validates CopySFA's H2D load overlap effect (Section~\ref{sec:transfer-overlap}). The kernel unit-test operating point follows Section~\ref{sec:setup}, with sequence length 65536; the baseline runs H2Dcopy and SFA serially as two kernels (\(T_\mathrm{copy}+T_\mathrm{SFA}\)). Figure~\ref{fig:copysfa-latency} gives the batch-size sweep from 1 to 32.

The results (Figure~\ref{fig:copysfa-latency}) show CopySFA yields positive gains in all configurations, with speedups of 1.03\(\times\)--1.27\(\times\). Defining \(\mathrm{OverlapRatio}=(T_\mathrm{copy}+T_\mathrm{SFA}-T_\mathrm{fused})/T_\mathrm{copy}\), the ratio reaches 90\%--98\% at batch-size \(\geq16\).

\subsection{End-to-End Full-Model Inference Experiments}\label{sec:e2e}

This section presents FlashOffloading's benefits in a real PD-disaggregated inference scenario, validating that prefill KV cache offloading and decode KV cache offloading can work together.

\subsubsection{Workload and Metrics}\label{sec:e2e-setup}

\textbf{Workload design.} A burst of concurrent requests is fired at the system: input lengths 64000--1000000 tokens, output length 1 (prefill only) or 8192 tokens, and the prefix hit rate set to the 90\% typical of mainstream scenarios. Three offloading configurations: none (both off), P-only (prefill offloading only), and P+D (dual offloading).

\textbf{Metrics.} Client side: P99 TE2E (from the burst start until the last request finishes decode), P99 TTFT (until the last request finishes prefill), avg TPOT, avg ITL, and avg input/output throughput (i.e., system-level prefill/decode throughput). Server side: the decode node's profiling kernel-latency breakdown, plus max-num-batched-tokens, per-die HBM usage, and the actual decode batch-size per DP rank.

\subsubsection{End-to-End Results}\label{sec:e2e-discussion}

\begin{figure}[t]
\centering
\includegraphics[width=0.9 \columnwidth]{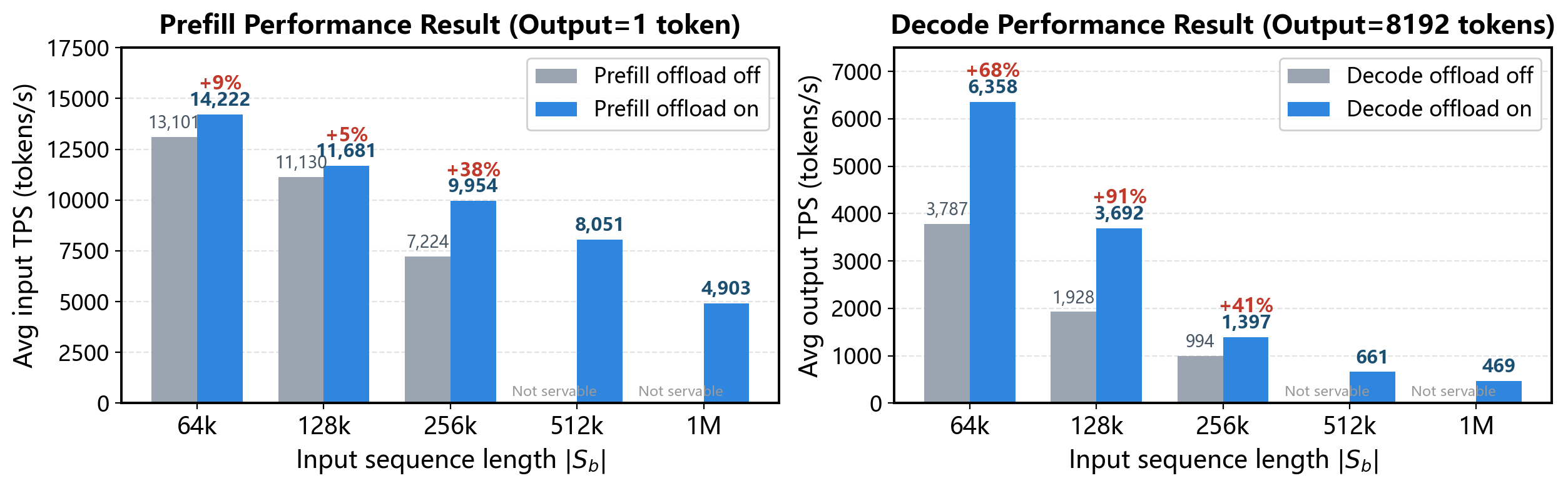}
\caption{Throughput comparison with and without offloading at different sequence lengths.}
\label{fig:e2e-tps}
\end{figure}

Table~\ref{tab:e2e} presents the end-to-end results: for each input length, we first compare prefill offloading off vs.~on with output length 1 (prefill only), then compare decode offloading off vs.~on with output length 8192 (prefill offloading on in both). Figure~\ref{fig:e2e-tps} visualizes the last two columns of Table~\ref{tab:e2e}.

\begin{table*}[t]
\centering
\caption{PD-disaggregated results on 32 Ascend910C (64 die), prefill DP2/TP16/EP32, decode DP16/TP2/EP32, prefix cache hit 90\%.}
\label{tab:e2e}
\scriptsize
\setlength{\tabcolsep}{1.5pt}
\resizebox{\textwidth}{!}{\begin{tabularx}{\textwidth}{@{}>{\hsize=0.85\hsize\centering\arraybackslash}X>{\hsize=0.85\hsize\centering\arraybackslash}X>{\centering\arraybackslash}X>{\hsize=0.85\hsize\centering\arraybackslash}X>{\centering\arraybackslash}X>{\centering\arraybackslash}X>{\centering\arraybackslash}X>{\hsize=1.5\hsize\centering\arraybackslash}X>{\centering\arraybackslash}X>{\centering\arraybackslash}X>{\centering\arraybackslash}X>{\centering\arraybackslash}X>{\centering\arraybackslash}X>{\centering\arraybackslash}X>{\centering\arraybackslash}X>{\centering\arraybackslash}X>{\centering\arraybackslash}X>{\centering\arraybackslash}X@{}}
\toprule
\multicolumn{2}{c}{Seq.\ Length} &  &  & \multicolumn{4}{c}{Prefill HBM usage} & \multicolumn{4}{c}{Decode HBM usage} & \multicolumn{4}{c}{Latency} & \multicolumn{2}{c}{Total Throughput} \\
\cmidrule(lr){1-2}\cmidrule(lr){5-8}\cmidrule(lr){9-12}\cmidrule(lr){13-16}\cmidrule(lr){17-18}
Input & Output & Offload \newline Mode & Client \newline BS & weight \newline (GB) & KV \newline (GB) & act. \newline (GB) & max \newline batched \newline toks. & weight \newline (GB) & KV \newline (GB) & KV \newline usage \newline (GB) & BS \newline per \newline NPU & P99 \newline TE2E \newline (s) & P99 \newline TTFT \newline (s) & Avg. \newline TPOT \newline (ms) & Avg. \newline ITL \newline (ms) & Input & Output \\
\midrule
64k & 1 & none & 192 & 18.86 & 36.75 & 1.67 & 16384 & 25.36 & 28.40 & -- & -- & 93.8 & 93.8 & -- & -- & 13101 & -- \\
64k & 1 & P-only & 192 & 18.86 & 9.04 & 13.26 & 131072 & 25.36 & 28.40 & -- & -- & 86.4 & 86.4 & -- & -- & 14222 & -- \\
64k & 8192 & P-only & 192 & 18.86 & 9.04 & 13.26 & 131072 & 25.36 & 28.40 & 24.36 & 4 & 459.3 & 320.7 & 16.9 & 63.6 & 3832 & 3787 \\
64k & 8192 & P+D & 192 & 18.86 & 9.04 & 13.26 & 131072 & 25.36 & 27.22 & 21.79 & 12 & 344.6 & 96.7 & 30.2 & 110.6 & 12707 & 6358 \\
\midrule
128k & 1 & none & 96 & 18.86 & 36.75 & 1.67 & 16384 & 25.36 & 28.40 & -- & -- & 110.4 & 110.4 & -- & -- & 11130 & -- \\
128k & 1 & P-only & 96 & 18.86 & 9.04 & 13.26 & 131072 & 25.36 & 28.40 & -- & -- & 105.2 & 105.2 & -- & -- & 11681 & -- \\
128k & 8192 & P-only & 96 & 18.86 & 9.04 & 13.26 & 131072 & 25.36 & 28.40 & 24.36 & 2 & 456.0 & 320.9 & 16.6 & 62.3 & 3829 & 1928 \\
128k & 8192 & P+D & 96 & 18.86 & 9.04 & 13.26 & 131072 & 25.36 & 27.22 & 12.96 & 6 & 333.2 & 122.8 & 26.0 & 96.1 & 10007 & 3692 \\
\midrule
256k & 1 & none & 32 & 18.86 & 36.75 & 1.67 & 16384 & 25.36 & 28.40 & -- & -- & 113.4 & 113.4 & -- & -- & 7224 & -- \\
256k & 1 & P-only & 32 & 18.86 & 9.04 & 13.26 & 131072 & 25.36 & 28.40 & -- & -- & 82.3 & 82.3 & -- & -- & 9954 & -- \\
256k & 8192 & P-only & 32 & 18.86 & 9.04 & 13.26 & 131072 & 25.36 & 28.40 & 24.36 & 1 & 321.7 & 191.1 & 16.1 & 60.4 & 4287 & 994 \\
256k & 8192 & P+D & 32 & 18.86 & 9.04 & 13.26 & 131072 & 25.36 & 27.22 & 8.27 & 2 & 317.7 & 128.9 & 22.9 & 84.3 & 6355 & 1397 \\
\midrule
512k & 1 & none & 32 & 18.86 & 36.75 & 1.67 & 16384 & 25.36 & 28.40 & \multicolumn{8}{c}{\tiny Not servable} \\
512k & 1 & P-only & 32 & 18.86 & 9.04 & 13.26 & 131072 & 25.36 & 28.40 & -- & -- & 203.5 & 203.5 & -- & -- & 8051 & -- \\
512k & 8192 & P-only & 32 & 18.86 & 9.04 & 13.26 & 131072 & 25.36 & 28.40 & \multicolumn{8}{c}{\tiny Not servable} \\
512k & 8192 & P+D & 32 & 18.86 & 9.04 & 13.26 & 131072 & 25.36 & 27.22 & 5.51 & 1 & 521.8 & 325.9 & 24.2 & 90.4 & -- & 661 \\
\midrule
1M & 1 & none & 32 & 18.86 & 36.75 & 1.67 & 16384 & 25.36 & 28.40 & \multicolumn{8}{c}{\tiny Not servable} \\
1M & 1 & P-only & 32 & 18.86 & 9.04 & 13.26 & 131072 & 25.36 & 28.40 & -- & -- & 652.6 & 652.6 & -- & -- & 4903 & -- \\
1M & 8192 & P-only & 32 & 18.86 & 9.04 & 13.26 & 131072 & 25.36 & 28.40 & \multicolumn{8}{c}{\tiny Not servable} \\
1M & 8192 & P+D & 32 & 18.86 & 9.04 & 13.26 & 131072 & 25.36 & 27.22 & 8.14 & 1 & 842.4 & 572.0 & 34.1 & 130.6 & 5594 & 469 \\
\bottomrule
\end{tabularx}}
\end{table*}

Table~\ref{tab:e2e} and Figure~\ref{fig:e2e-tps} yield six conclusions:

\begin{itemize}
\tightlist
\item
  \textbf{Ultra-long sequences become servable}: 512k/1M sequences need 48.76/97.52 GB KV cache per request, exceeding D-node capacity even for one request; only P+D serves \(\geq\)512k.
\item
  \textbf{Long-sequence prefill improvement}: at output length 1, P-only lowers P99 TTFT by 7.9\%/4.7\%/27.4\% (64k/128k/256k) and raises input TPS by 8.6\%/5.0\%/37.8\% --- longer inputs gain more.
\item
  \textbf{Long-sequence decode improvement}: at output length 8192, P+D raises the actual decode batch-size 3\(\times\)/3\(\times\)/2\(\times\) over P-only (64k/128k/256k) while TPOT degrades 78.7\%/56.6\%/42.2\% (\(\sim\)30 ms level); output TPS rises 67.8\%/91.5\%/40.5\%.
\item
  \textbf{P-node HBM}: prefill offloading cuts total use 60.0\(\rightarrow\)50.5 GB: KV preallocation 36.75\(\rightarrow\)9.04 GB (4 rolling buffers; max servable sequence \(\sim\)368k\(\rightarrow\)\(\sim\)1.6M), activations 1.67\(\rightarrow\)13.26 GB, max-num-batched-tokens 16384\(\rightarrow\)131072 (8\(\times\)).
\item
  \textbf{D-node HBM}: decode offloading cuts total use 63.7\(\rightarrow\)57.3 GB; per-request residency at 64k falls 6.09\(\rightarrow\)1.78 GB (16k hot-cache budget), so 4\(\rightarrow\)15 requests fit per DP rank (12 in production to control TPOT).
\item
  \textbf{Under P-only, TTFT at output 8192 exceeds output 1} (64k: 320.7 s vs.~86.4 s): the D node paces PD disaggregation, and longer outputs let decode block prefill.
\end{itemize}

\subsubsection{Decode End-to-End Latency Breakdown}\label{sec:e2e-results}

\begin{figure}[t]
\centering
\includegraphics[width=0.75 \columnwidth]{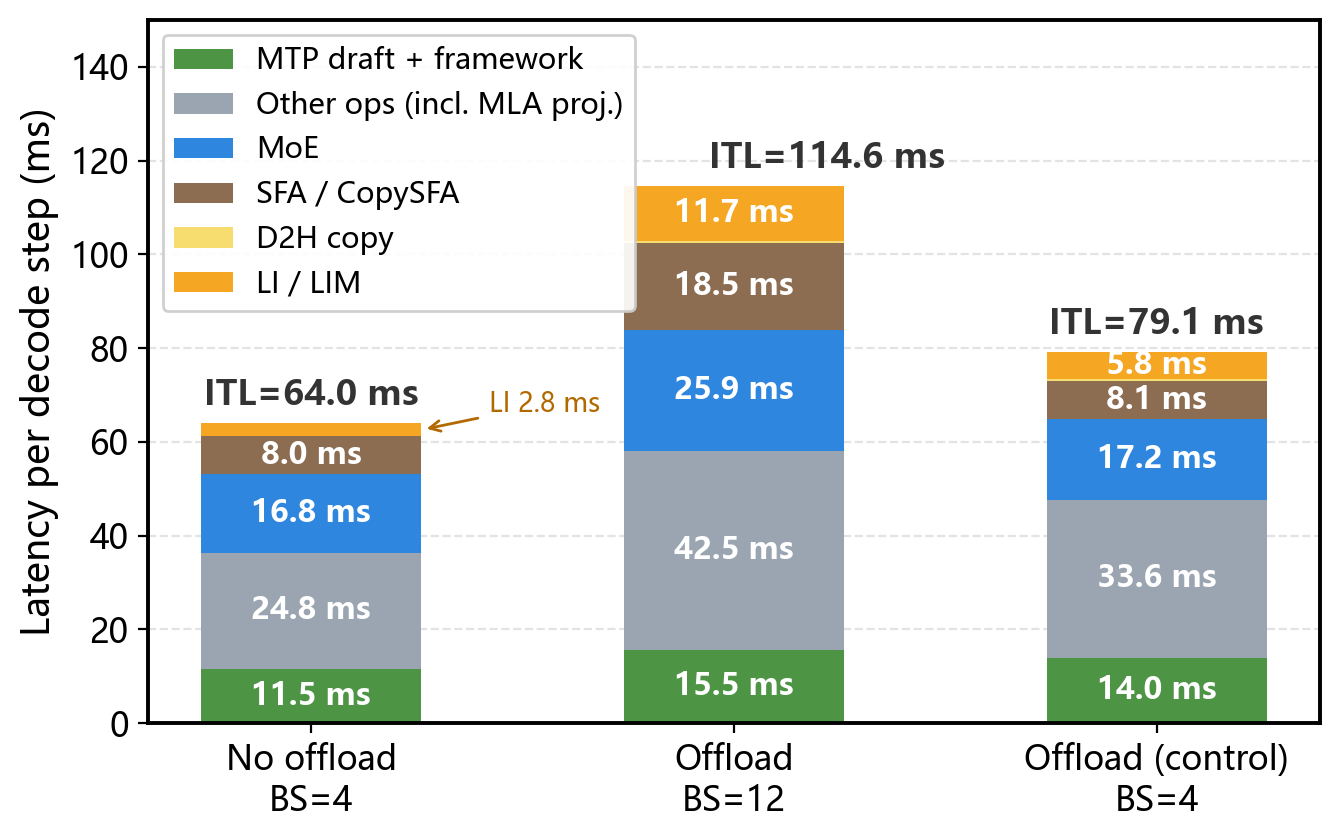}
\caption{Kernel-level breakdown of one decode step at 64k input.}
\label{fig:breakdown}
\end{figure}

This section presents each kernel's share of ITL in steady-state decode to reveal the source of the TPOT degradation. Figure~\ref{fig:breakdown} shows three groups at 64k input: no offloading with BS=4 (baseline), dual offloading with BS=12 (production operating point), and dual offloading with BS=4 (control group, same compute as the baseline).

Figure~\ref{fig:breakdown} yields two conclusions:

\begin{itemize}
\tightlist
\item
  \textbf{Latency control group (dual offloading BS=4 vs.~no offloading BS=4): offloading's inherent latency increment concentrates in the two optimized components.} ITL +15.2 ms (+23.7\%): LIM +3.0 ms over LI (steps \stepnum{1}--\stepnum{4} management; matches Section~\ref{sec:index-latency}'s 21 layers \(\times\) \(\sim\)144\ensuremath{\mu}s); CopySFA only +0.1 ms over SFA (step-\stepnum{5} load almost fully overlapped; 90\%+ ratio, Section~\ref{sec:overlap-effect}); MoE/SFA compute identical, latency flat. ``Other kernels'' (e.g., MLA projection) +8.8 ms, from MoE-dispatch synchronization: offloading makes DP ranks' miss distributions differ, so early-finishing ranks idle at the sync point --- deployment-level imbalance, not inherent overhead.
\item
  \textbf{Production group (dual offloading BS=12 vs.~no offloading BS=4): ITL grows far less than batch-size.} ITL +50.6 ms (+79\%): LIM (11.7 ms) and CopySFA (18.5 ms) grow with the \(\mathrm{BS} \times\) miss-union load, MoE sublinearly to 25.9 ms, sync wait 42.5 ms. TPOT 17.0\(\rightarrow\)30.2 ms (+77\%), still acceptable; batch-size \(\times\)3 with ITL \(\times\)1.79, so steady-state decode throughput improves \(\approx\)\(\times\)1.7 (Section~\ref{sec:e2e-discussion}).
\end{itemize}

\section{Discussion}\label{sec:discussion}

\textbf{Implications for model design.} To overcome the KV cache HBM capacity wall, model architectures are evolving along three routes: DSA (GLM-5, DeepSeek-V3.2 \cite{ref15,ref16}), extreme compression (e.g., DeepSeek-v4.1), and linear attention (e.g., Kimi K3). The latter two cut capacity but discard information and impair capability; DSA keeps capacity and only exploits attention's inherent sparsity to dynamically ignore some tokens, with the least capability loss. Without low-latency offloading, DSA's capacity cost would push models toward lossy routes; FlashOffloading hosts DSA's full KV history in host DRAM with controlled latency, making the DSA route engineering-viable and letting models evolve toward minimal capability loss.

\textbf{Implications for accelerator hardware design.} FlashOffloading shows that next-generation NPUs/GPUs can ship with less HBM to cut cost, offloading both prefill and decode KV cache to the prefix-cache DRAM layer that inference systems already have.

\section{Conclusion}\label{sec:conclusion}

This paper presents FlashOffloading, a full KV cache offloading architecture for PD-disaggregated long-context inference: host DRAM holds the authoritative KV replica, while residency and transfer granularity switch by stage. Prefill pipelines a bounded four-layer buffer pool; decode decouples cache-management latency from the cache budget (cache\_slots, LIM) and overlaps miss loads inside sparse attention (CopySFA). On Ascend910C, cache management adds only hundred-microsecond latency and up to 98\% of miss loads are overlapped; end to end, decode throughput at 64K improves by 1.68\(\times\), and 512K/1M sequences become servable --- all without changing the model-defined attention set: offloading changes only data placement, not computed results.
\clearpage
\nocite{*}
\bibliography{references}
\bibliographystyle{mlsys2025}
\end{document}